\documentclass[10pt, a4paper]{article}

\usepackage[]{lrec-coling2024} 

\usepackage{comment}
\usepackage{graphicx}
\usepackage{booktabs}
\usepackage{xcolor}
\usepackage{soul}
\usepackage{multirow}
\usepackage{subfigure}
\usepackage{amssymb} 
\usepackage{rotating}
\usepackage{xurl}
\usepackage{makecell}

\title{SPACE-IDEAS: A Dataset for Salient Information Detection in Space Innovation}

\name{Andrés García-Silva, Cristian Berrío, José Manuel Gómez-Pérez} 
\address{Expert.ai, Language Technology Research Laboratory,\\ 
Poeta Joan Maragall 3-5, 28020 Madrid, Spain \\
agarcia@expert.ai, cberrio@expert.ai, jmgomez@expert.ai} 

\abstract{
Detecting salient parts in text using natural language processing has been widely used to mitigate the effects of information overflow. Nevertheless, most of the datasets available for this task are derived mainly from academic publications. We introduce SPACE-IDEAS, a dataset for salient information detection from innovation ideas related to the Space domain. The text in SPACE-IDEAS varies greatly and includes informal, technical, academic and business-oriented writing styles. In addition to a manually annotated dataset we release an extended version that is annotated using a large generative language model. We train different sentence and sequential sentence classifiers, and show that the automatically annotated dataset can be leveraged using multitask learning to train better classifiers.  
 \\ \newline  }

\begin{document}

\maketitleabstract

\section{Introduction}
In recent years, the number of research and innovation content has grown substantially \cite{krenn2022predicting}. Open source publications, digital publications, and preprints servers have contributed to this growth. Detecting salient fragments of text contributes to mitigate information overload, helping readers to focus on the most important parts.

Detecting salient parts in text has been tackled as a sequential sentence classification task, where sentences are categorized into their respective roles considering that the label of each sentence is related to the surrounding sentences \citep{Jin2018HierarchicalNN}.  
Typically, sequential sentence classifiers are trained using supervised learning \citep{Gonalves2019ADL,Jin2018HierarchicalNN, Yamada2020SequentialSC,cohan2019pretrained}. 

Annotated datasets to train sequential sentence classifiers are mostly focused on the scholarly communication domain. For example the CSAbtruct dataset \citep{cohan2019pretrained} includes abstracts from computer science publications, the Scim dataset \citep{Fok2022ScimIF} contains full publications from NLP conferences, and PubMed RCT \cite{dernoncourt2017pubmed} and NICTA \cite{Kim2011AutomaticCO} are centered on the biomedical domain.  

In this paper we introduce SPACE-IDEAS: a Dataset for Salient Information Detection in Space Innovation. SPACE-IDEAS is manually annotated following a methodology that ensures high quality annotations. Additionally, we  release SPACE-IDEAS+, a larger dataset annotated with assistance of OpenAI's gpt-3.5-turbo. We use the same set of instructions and examples provided to human annotators when prompting the generative language model. The percentage of agreement between gpt-3.5-turbo annotations and gold annotations is reasonably close to the initial agreement between human annotators as shown in our quality analysis. 

SPACE-IDEAS differs in several aspects to existing datasets. It covers the Space domain, which was not previously included in any dataset. Moreover the text comes from public ideas in the Open Space Innovation Platform OSIP\footnote{\url{https://ideas.esa.int}}. Although ideas may look similar to abstracts since they are both brief overviews of a longer document, they are very different. Academic abstracts summarize completed research, adhering to academic conventions and catering to formal writing. In contrast, ideas pitch a project or innovation not implemented yet, often with a non formal nor academic writing style. 

Along the dataset we contribute a baseline  classifier that we trained on top of a pre-trained language model using multitask learning. We rely on the approach presented in \citep{cohan2019pretrained} for sequential sentence classification since it allows us to easily plug in, fine-tune, and test state of the art transformers. We test different transfer learning techniques \cite{hedderich2021survey} to leverage training data in SPACE-IDEAS and SPACE-IDEAS+ datasets. The datasets and code to reproduce our experiments are publicly available.\footnote{\url{https://github.com/expertailab/SPACE-IDEAS}}

\section{Related work}\label{sec:soa}
Datasets for role sentence classification (table \ref{tab:datasets}) contain academic abstracts and full papers covering domains such as biomedicine (PMD-RCT \citep{dernoncourt2017pubmed} and NICTA-PIBOSO \cite{Kim2011AutomaticCO}), computer science (CSAbstruct \citep{cohan2019pretrained}, CS-Abstracts \cite{Gonalves2019ADL}, scim \cite{Fok2022ScimIF}, and Dr Inventor \cite{fisas-etal-2015-discoursive}). Others are multidisciplinary, like Emerald 100k \cite{Stead2019Emerald1A}, MAZEA \cite{dayrell-etal-2012-rhetorical}, and ART/CoreSC \cite{liakata-etal-2010-corpora}.


A common architecture for sequential sentence classification consists of hierarchical encoders of words and sentences to contextualize sentence representations \citep{Jin2018HierarchicalNN,Shang2021ASD,Brack2021CrossDomainML}, and an output layer to predict the labels. Others, like \citet{cohan2019pretrained} use BERT to leverage contextualized representations of all words in all sentences. The output layer is often a SoftMax classifier \citep{cohan2019pretrained, Gonalves2019ADL} or a conditional random field (CRF) layer \cite{dernoncourt2017pubmed,  Yamada2020SequentialSC} to consider the interdependence between labels. 

\section{SPACE-IDEAS Dataset}
\begin{figure}[t!]
    \centering
    \resizebox{.95\linewidth}{!}{
        \includegraphics{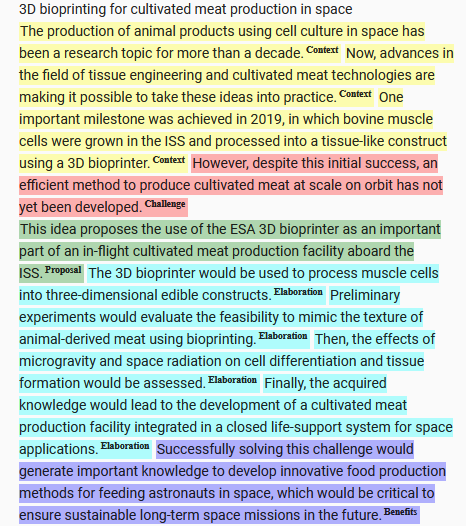}
    }
    \caption{Example annotations of salient sentences in an Idea submitted to OSIP}
    \label{fig:labeled_idea}
\end{figure}

In collaboration with the OSIP team, we have identified the following roles that sentences serve in ideas: Challenge, Proposal, Elaboration, Benefits, and Context. Typically, an idea addresses a challenge in a particular context and proposes a solution which is the core of the idea. The solution is then elaborated and its benefits made explicit (see fig. \ref{fig:labeled_idea}).

To create the dataset, we gather a random sample of 176 ideas from the Open Space Innovation Platform (OSIP) that are marked as not confidential by their authors. The dataset contains 1733 sentences and 49420 words. On average an idea has 9.8 sentences with a standard deviation of 3.8. 

The annotation process has two stages. In the first stage, each annotator labels a set of ideas. We make sure each idea is annotated by two annotators. In the second stage, we identify the disagreements among annotators and arrange meetings between pairs of annotators so that they can agree on the final annotations.

We engage six annotators, all of whom are university graduates with prior experience in annotation processes. We hand each annotator the annotation guidelines\footnote{\url{https://github.com/expertailab/SPACE-IDEAS/blob/master/AnnotationGuidelines.pdf}} that define the goals of the annotation process, the labels to annotate the sentences, and three exemplary ideas completely annotated. We meet with each annotator to discuss the annotation guidelines, solve any doubt, and explain how to use label studio\footnote{\url{https://labelstud.io/}}, the tool supporting the annotation process. In the first annotation stage each annotator labels 60 ideas approximately


After the first stage, the percentage of agreement among annotators is 0.65 and Cohen\textquotesingle s kappa coefficient, which measures inter-rater agreement considering the possibility of the agreement occurring by chance, is 0.56. Disagreements were settled in the second stage. In other words, the dataset's final annotations are the result of agreement between two annotators. In total each annotator spent 8 hours approximately annotating ideas and 4 additional hours in bilateral meetings solving disagreements. 

\begin{table*}[htbp]
  \centering
  \resizebox{\linewidth}{!}{ 
  \setlength\tabcolsep{2pt}
    \begin{tabular}{cccccl}
    \toprule
    \textbf{Dataset} & \multicolumn{1}{c}{\textbf{Domain}} & \textbf{Doc Type} & \textbf{Instances} & \textbf{Sentences} & \multicolumn{1}{c}{\textbf{Labels}} \\
    \midrule
    SPACE-IDEAS & Space & Idea  & 176   & 1733  & Challenge (12\%), Proposal (14\%), Elaboration (32\%), Benefits (10\%), Context (33\%) \\
    SPACE-IDEAS+ & Space & Idea  & 1020  & 9806  & Challenge (17\%), Proposal (20\%), Elaboration (16\%), Benefits (16\%), Context (31\%) \\
    CSAbstruct & Comp. Science & Abstract & 2189  & 4730  & Background (33\%), Objective (12\%), Method (32\%), Result (21\%), Other (3\%) \\
    Scim  & NLP   & Full paper & 3051  & 606K  & Objective (4\%), Method (14\%), Result (4\%), Other (10\%), Abstain (67\%) \\
    PMD-RCT & Biomedical & Abstract & 20000 & 2.3M  & Background (33\%), Objective (12\%), Method (32\%), Result (21\%), Other (3\%) \\
    NICTA-PIBOSO & Biomedical & Abstract & 1000  & 10379 & \makecell[tl]{Background (25\%), Intervention (7\%), Study (2\%), Population (8\%), Outcome (43\%),\\Other (15\%)} \\
    CS-Abstracts & Comp. Science & Abstract & 654   & 4730  & Background, Objective, Methods, Results, Conclusions \\
    Emerald 100k & \makecell[t]{Management, Engin.,\\Information Science} & Abstract & 103457 & 1050397 & \makecell[tl]{Purpose (19\%), Design/methodology/approach (21\%), Findings (26\%),\\Originality/value (18\%), Social implications (0.002\%), Practical implications (9\%),\\Research limitations/implications (7\%)} \\
    MAZEA & \makecell[t]{Physics, Engin.,\\Life and Health Sci's} & Abstract & 1335  & 13477 & Background, Gap, Purpose, Method, Result, Conclusion \\
    Dr. Inventor & Comp. Graphics & Full paper & 40    & 10789 & \makecell[tl]{Background (16.32\%), Approach (46.70\%), Challenge (3.25\%), Challenge\_Goal (0.84\%), \\Challenge\_Hypotesis (0.06\%),  Outcome (10.89\%), Outcome\_Contribution (2.03\%), \\Future Work (1.26\%), Unspecified  (7.04\%), Sentence (11.61\%)} \\
    ART/CoreSC & \makecell[t]{Chemistry,\\Comp. Ling.} & Full paper & 225   & 35040 & \makecell[tl]{Background, Motivation, Goal, Hypothesis, Object, Model, Method, Experiment,\\Result, Observation, Conclusion} \\
    \bottomrule
    \end{tabular}%
    }
  \caption{Characterization of datasets for sentence classification including SPACE-IDEAS and SPACE-IDEAS+. In round brackets the percentage of each label in the dataset as published.}
  \label{tab:datasets}%
\end{table*}%

\subsection{SPACE-IDEAS+}\label{osip+}
SPACE-IDEAS contains high-quality annotations on a reduced number of ideas. \citet{Brack2021CrossDomainML} show that using  transfer learning from semantically related tasks and datasets benefits sequential sentence classifiers when limited training data is available. Rather than using related datasets, we annotate a larger set of ideas using a generative large language model.
We prompt\footnote{Prompt example: \url{https://github.com/expertailab/SPACE-IDEAS/blob/master/chatgpt_prompt.md}} the gpt-3.5-turbo model \footnote{\url{https://platform.openai.com/docs/api-reference/chat}} with the annotation guidelines that we provide to the human annotators, including four examples of ideas fully annotated. The prompt instructs the model to annotate each sentence by appending a label at the end of each sentence. 
Then we ask the model to annotate each idea following the guidelines and considering the examples provided. The final annotated dataset, that we call SPACE-IDEAS+, contains all publicly available ideas, totalling 1020 ideas and 9806 sentences. 

We assess the quality of the generated dataset by measuring the agreement between GPT annotations and the human annotations we collected in SPACE-IDEAS. The percentage of agreement reached in the intersection of both datasets, which includes all ideas in SPACE-IDEAS, is 0.5. While the agreement is lower than what human annotators achieved in the initial annotation stage, it is significantly higher than 
the random annotation probability of 0.2 given 5 possible labels. Moreover, the agreement exceeds the probability of selecting the majority label which stands at 0.33, representing the prevalence of the 'Context' label in the SPACE-IDEAS dataset. Notably, the agreement level remains within a reasonable range compared to human agreement (0.65), which serves as the upper bound of agreement.

\subsection{Fields and label Distribution}
To identify the knowledge fields in the dataset, we train a text classifier relying on a RoBERTa model \cite{liu2019roberta} and a public corpus gathered from the NASA Technical report server (NTRS)\footnote{\url{https://ntrs.nasa.gov/}}, where project descriptions are annotated with categories describing the knowledge fields. These categories are described in the NASA Scope and Subject Category guide\footnote{\url{https://ntrs.nasa.gov/citations/20000025197}}, which is also publicly available.

\begin{figure}[ht!]
    \centering
    \resizebox{\linewidth}{!}{        \includegraphics{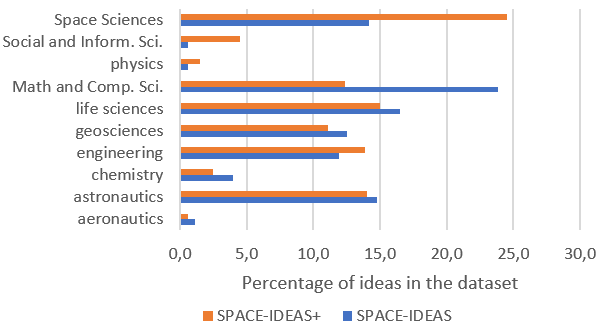}
    }
    \caption{Distribution of knowledge fields in the SPACE-IDEAS datasets}
    \label{fig:fielddistribution}
\end{figure}

In fig. \ref{fig:fielddistribution} we show the  distribution of knowledge fields in SPACE-IDEAS  and SPACE-IDEAS+. Both datasets have similar distribution, with exception of the Space Sciences field that is more represented in SPACE-IDEAS+ and Math and Computer Science that is more represented in SPACE-IDEAS. Considering the label distribution shown in table \ref{tab:datasets}, we can observe that both datasets exhibit similar label distributions, with differences of less than 6\% for most labels, except for the 'Elaboration' label, where the difference rises to 16\%. Despite such difference, we show in our experiments that data in SPACE-IDEAS+ contributes positively to learn better classifiers.

\subsection{Relation to existing datasets}
In contrast to most of available datasets for sequential sentence classification centered in academic publications, SPACE-IDEAS contains innovative ideas in the space domain (see table \ref{tab:datasets}). Ideas can be written using informal, technical or business-oriented language. Such variety of writing styles is an additional challenge that our dataset introduces. Moreover, SPACE-IDEAS includes four  knowledge fields within the space domain (Space Sciences, Geosciences, Astronautics, Aeronautics) that are not covered in any dataset. Considering the number of documents, while SPACE-IDEAS dataset is relatively small, it is still larger than three other datasets. Note that SPACE-IDEAS+ includes all the public ideas in OSIP at the time of collection.

\section{Detecting salient information}
Salient information detection is a multi-class classification task where a sentence is only assigned one label. Given an idea description $D$ consisting of $k$ sentences, the task is to assign a label $l_{i}$ from a set of $n$ possible labels to each sentence in $D$. The set of labels in SPACE-IDEAS is $L=$ \textit{\{Challenge, Proposal, Elaboration, Benefits, Context\}}. We propose the following baselines to address this task.

\textbf{Single-sentence classifier.} Following the usual fine-tuning approach for text classification, we use as aggregate sentence representation the final output vector $C \in R^H$ for the first input token, where H is the hidden size in the transformer. $C$ is connected to a softmax classification layer $W\in^{K \times H}$, where K is the number of classes and cross-entropy is the loss function. 

\textbf{Sequential Sentence Classification.} While single-sentence classification does not capture the relations between sentences in the text, sequential sentence classification assigns a label to each sentence simultaneously, making a better use of context. The approach we are using, as outlined by \citet{cohan2019pretrained}, involves a transformer encoder that jointly encodes and contextualizes all sentences. Sentences in an idea are concatenated using the separator token and fed into the transformer. We use as the sentence representation the output vector $S_{i} \in R^H$ corresponding to each separator token. Each output vector $S_{i}$ is then connected to an output classification layer. 


\textbf{Transfer learning.} In addition, we use transfer learning \cite{ruder-etal-2019-transfer} to leverage data from SPACE-IDEAS and SPACE-IDEAS+. We apply sequential transfer learning by first fine-tuning a pre-trained  transformer model on the larger dataset, SPACE-IDEAS+, and then fine-tune the same transformer model using SPACE-IDEAS. 
We also use multi-task learning where several classifiers are learned simultaneously. We add different classification heads, one per dataset, on top of a pre-trained transformer that acts as shared model. \citet{Brack2021CrossDomainML} show that for sequential sentence classification multi-task learning is more effective in low data scenarios than sequential transfer learning.


\section{Experiments}
We use RoBERTa large \cite{liu2019roberta} as encoder in the classifiers. We use the output vector for token <s> as the sequence aggregated representation in single sentence classifiers, and output vectors for separation tokens </s> as sentence representations in sequential sentence classifiers. We hold out 20\% of the SPACE-IDEAS dataset for testing. From the remaining 80\%, we use 80\% for training and 20\% for validation. When we use SPACE-IDEAS+, we train on the whole dataset and evaluate in the test set of SPACE-IDEAS. As evaluation metric, we use micro F1-score and span-F1 \cite{Yamada2020SequentialSC}, which rather than evaluating labeling at the sentence level, evaluates whether a span of contiguous sentences is labeled correctly. 

We train single-sentence classifiers with 2e-5 learning rate,\footnote{Learning rate is adjusted using the validation set.} batch size 2, and gradient accumulation. When we use additional context, we append the whole idea description to the input sentence using the </s> token. To train sequential-sentence classifiers,\footnote{\url{https://github.com/UrszulaCzerwinska/sequential_sentence_classification/tree/allennlp2}} we use a learning rate of 1e-5, a batch size of 1, and gradient accumulation. We train for a maximum of 20 epochs, using early stopping with a patience of 3 epochs. 
We train each classifier three times and report the average metrics in table \ref{tab:experiments}. 

\noindent
\begin{table}[t!]
  \centering
   \resizebox{\linewidth}{!}{  
    \begin{tabular}{lllrr}
    \toprule
    Classifier & Context & Dataset & \multicolumn{1}{l}{F1} & \multicolumn{1}{l}{Span-F1} \\
    \midrule
    Sent. & & SPACE-IDEAS  &   63.5    & 39.0  \\
    Sent. & \checkmark & SPACE-IDEAS  & \underline{71.1} &  \underline{47.9}    \\    
    Seq. Sent. & & SPACE-IDEAS  &   68.5    &   44.0     \\
    Sent. & & SPACE-IDEAS+ &    57.5   &   39.0     \\
    Sent. & \checkmark & SPACE-IDEAS+ &    56.6   &   38.0   \\
    Seq. Sent. & & SPACE-IDEAS+ &   58.2    &   39.6    \\
    \midrule
    \multicolumn{5}{c}{Sequential Transfer} \\
    \midrule
    Sent. & \checkmark & SPACE-IDEAS+ SPACE-IDEAS &   70.2    &    44.6    \\
    Seq. Sent. & & SPACE-IDEAS+ SPACE-IDEAS &   \underline{70.7}    &   \underline{46.7}      \\
    \midrule
    \multicolumn{5}{c}{Multi-Task Transfer} \\
    \midrule
    Sent. & \checkmark & SPACE-IDEAS+ SPACE-IDEAS &   68.5    &    45.3    \\
    Seq. Sent. & & SPACE-IDEAS+ SPACE-IDEAS &   \textbf{72.6}    &    \textbf{50.0}    \\
    \bottomrule
    \end{tabular}%

  }
  \caption{Evaluation results of different classifiers trained on the SPACE-IDEAS and SPACE-IDEAS+ datasets}  
  \label{tab:experiments}
\end{table}%

The best classifier was trained using SPACE-IDEAS+ and SPACE-IDEAS in a multi-task learning objective, reaching 72.6 F1-score and 50.0 span-F1. The confusion matrix for such model in table \ref{tab:confmat} shows that the classifier learns to predict labels Context, Proposal and Benefits with high precision, and to a lesser extent Challenge and Elaboration, which are confused with Context mainly. In addition, the classifier exhibits high recall for labels Challenge, Proposal and Elaboration. 

\begin{table}[t!]
  \centering
  \resizebox{\linewidth}{!}{  
    \begin{tabular}{crccccc}
    \toprule
          & \multicolumn{1}{r}{} & \multicolumn{5}{c}{Predicted Value} \\
          & \multicolumn{1}{r}{} & \multicolumn{1}{r}{Challenge} & \multicolumn{1}{r}{Context} & \multicolumn{1}{r}{Proposal} & \multicolumn{1}{r}{Benefits} & \multicolumn{1}{r}{Elaboration} \\
\cmidrule{3-7}    \multicolumn{1}{r}{\multirow{5}[2]{*}{\begin{sideways}Actual Value\end{sideways}}} & Challenge & 34    & 4     & 1     & 0     & 8 \\
          & Context & 15    & 70    & 3     & 1     & 20 \\
          & Proposal & 0     & 4     & 41    & 0     & 6 \\
          & Benefits & 0     & 0     & 1     & 22    & 10 \\
          & Elaboration & 4     & 8     & 6     & 3     & 90 \\
    \bottomrule
    \end{tabular}%
    }
    \caption{Confusion matrix of the sequential sentence classifier trained using multi-task learning.}
  \label{tab:confmat}%
\end{table}%

Surprisingly the single-sentence classifier using only SPACE-IDEAS, where we append the context to the input sentence, is the second best classifier. Such classifier improves over the single sentence classifier without using context on 8.9 points, and the sequential sentence classifier on 3.9 points. Moreover, sequential sentence classifiers improve over single sentence classifiers when more training data, combining both datasets, is available.

\section{Conclusion}
We present the SPACE-IDEAS dataset, which consists of public ideas submitted to the Open Space Innovation Platform hosted by the European Space Agency, where sentences are manually annotated with labels indicating their role in the text. We also release a larger dataset (SPACE-IDEAS+) automatically annotated using a generative approach. SPACE-IDEAS is the first dataset for sequential sentence classification covering knowledge fields related to the space domain not previously covered in any resource. We show through experimentation that leveraging both datasets to train classifiers in a multi-task setting leads to higher performance. A sequential sentence classifier trained on SPACE-IDEAS is currently deployed to highlight salient parts in the text of ideas submitted to OSIP.

\section{Ethics Considerations} General ethics consideration applies to classifiers trained and deployed using the SPACE-IDEAS datasets including transparency and accountability. Transparency is an issue if the classifier does not provide explanations about the label assigned to a group of sentences, which is the case of transformer-based  classifiers as the ones presented in this paper. Transparency can be enhanced with good user documentation and the integration of explainability techniques. Moreover, if a decision making process relies on the labels assigned by classifier, then in case of an incorrect decision there might be the question of who is accountable. Accountability can be improved with the definition of responsibilities, transparency reports including classifier performance, and good documentation. Considering privacy, in SPACE-IDEAS we only include text from ideas explicitly marked by the authors as non-confidential.

\section*{Acknowledgements}
We gratefully acknowledge the guidance and support of Moritz Fontaine and Charles-Antoine Poncet from ESA in shaping our project and contributing to its successful completion. We also extend our thanks to the expert.ai staff for their assistance during the annotation process.

\section{Bibliographical References}\label{sec:reference}

\bibliographystyle{lrec-coling2024-natbib}
\bibliography{lrec-coling2024-example}

\label{lr:ref}
\bibliographystylelanguageresource{lrec-coling2024-natbib}
\bibliographylanguageresource{languageresource}

\end{document}